\documentclass[review,authoryear]{elsarticle}
\usepackage{amssymb}
\journal{XXXXX}

\usepackage[utf8]{inputenc}
\usepackage{lineno}
\usepackage{hyperref}
\usepackage{amsmath}
\usepackage[toc,page]{appendix}
\usepackage{color,soul}
\usepackage{url}
\usepackage{caption}
\usepackage{subcaption}
\usepackage{fancyvrb}
\usepackage{listings}
\usepackage{hhline}
\usepackage{float}
\usepackage{adjustbox}
\usepackage{amsmath}
\usepackage{float}
\usepackage{multirow}
\usepackage[table,xcdraw]{xcolor}
\usepackage{todonotes}
\usepackage{algorithm}
\usepackage{algpseudocode}
\usepackage{footmisc}
\usepackage[normalem]{ulem}
\usepackage{placeins}
\usepackage{fixltx2e}
\usepackage{csquotes}
\usepackage{siunitx}

\modulolinenumbers[5]

\journal{Journal of \LaTeX\ Templates}

\sethlcolor{yellow}

\renewcommand{\sout}[1]{}

\begin{document}
\begin{frontmatter}

\title{Bridging Natural Language Processing and Psycholinguistics: computationally grounded semantic similarity datasets for Basque and Spanish}

\author{Josu Goikoetxea, Miren Arantzeta, Itziar SanMartin\footnote[2]{Correspondence authors: \textsf{josu.goikoetxea@ehu.eus, miren.arantzeta@ehu.eus, itziar.sanmartin@ehu.eus}}}

\begin{abstract}



We present a computationally-grounded word similarity dataset based on two well-known Natural Language Processing resources; text corpora and knowledge bases. This dataset aims to fulfil a gap in psycholinguistic research by providing a variety of quantifications of semantic similarity in an extensive set of noun pairs controlled by variables that play a significant role in lexical processing. The dataset creation has consisted in three steps, 1) computing four key psycholinguistic features for each noun; concreteness, frequency, semantic and phonological neighbourhood density; 2) pairing nouns across these four variables; 3) for each noun pair, assigning three types of word similarity measurements, computed out of text, Wordnet and hybrid embeddings. The present dataset includes noun pairs' information in Basque and European Spanish, but further work intends to extend it to more languages.

\end{abstract}


\end{frontmatter}


\newcommand\tab[1][1.2cm]{\hspace*{#1}}

\section{Introduction}
\label{sec:intro}

Semantic similarity is a measure of distance between items based on how similar their meanings are. That is, it is a type of relationship based on shared characteristics between concepts. For instance, 'cat' has more remarkable semantic similarity with 'tiger' than 'rhinoceros'. In contrast, semantic relatedness, frequently used interchangeably with semantic similarity, denotes any relation between concepts, not necessarily taxonomical. For instance, 'cat' is related to 'tiger', 'milk', and 'veterinary'. In short, semantic similarity offers a metric of categorical semantic relations, whereas semantic relatedness is closer to depicting associative relations. By now, we will use these two concepts interchangeably and show how the metrics used in the present dataset constitute independent proxies of different types of semantic relations.

Semantic relations impact the underlying capability of many cognitive processes, such as memory recall, mental imagery or language comprehension. They have been the focus of extensive research in Natural Language Processing (NLP) and psycholinguistics. There are computational tools to gauge the strength of semantic connections between concepts. Two important semantic resources in NLP are text corpora and knowledge bases (KB), both of them useful in cognitive experiments involving semantic similarity. In the case of KBs, they provide a quantitative and structured framework for representing words' meanings. 

This work presents two automatically calculated semantic similarity datasets restricted to nouns, to provide useful material to build up psycholinguistic experiments; one in Basque and the other in Spanish. The similarity measurements of noun pairs have been computed using vectors or embeddings built up in text corpora and KB. In order to capture different nuances of semantic similarity between noun pairs, in this work we employed three types of word representations; text embeddings, KB-based embeddings and hybrid embeddings (see \ref{sec:emb}). The selection of noun pairs has been conducted by controlling for word frequency, concreteness, and the number of semantic and phonologic neighbours due to the widely proven effect of these variables in language processing. Controlling for these linguistic features allows leveraging crucial lexical properties within pairs of nouns beyond semantic similarity. It broadens the usability of the dataset and the interpretation of the results.

In the following lines, we will describe the procedures that have been followed in each stage of the construction of the dataset, as well as the characteristics of the final dataset.


\section{Related work}
\label{sec:relwork}

Psycholinguistics and NLP have been closely linked for a long time. In semantic similarity computation, two methods have been mainly used; KB-based and embedding-based. The former treats KBs as graphs and exploit their complete structural information, so that similarity measurements are based on the KB taxonomy. The latter encodes word meaning in a numeric vector or embedding in a Euclidean space following the Distributional Hypothesis \citep{harris1954distributional}. Thus, embeddings latently encode the semantic (and also syntactic) features, and therefore, they offer comparable representations for words. 


Embedding measurements have revolutionized the field of NLP, increasing model accuracy significantly in practically every task. Further, they have overcome KB-based ones, proving to be more robust resources than the latter in word similarity tasks \citep{lastra2019reproducible}. 

In recent years, several works have given meaningful insights into the correspondence between word-embedding features and human cognition. Despite the misconceptions in understanding word embedding from a cognitive perspective \citep{gunther2019vector}, they remain the most meaningful proxy for human semantic representations. Thus, the performance of embeddings in the semantic similarity tasks has gained attention lately in combination with a variety of experimental methods, such as semantic priming \citep{ettinger2016evaluating,auguste2017evaluation,hollenstein2019cognival}, brain imaging \citep{sogaard2016evaluating,abnar2017experiential, ruan2016exploring, jain2018incorporating, rodrigues2018predicting, toneva2019interpreting} or eye-tracking \citep{luke2018provo,cop2017presenting,hollenstein2019cognival} (see also \citep{bakarov2018survey}). 

\cite{mandera2017explaining} proved that word vectors successfully explain semantic priming data, stating that they equal or overperform human-rated association datasets or feature norms. \cite{salicchi2021looking} showed a strong correlation between similarity measurements of contextual and non-contextual embeddings and two English corpora annotated by eye-tracking measures. \cite{hayes2021looking} explored the relationship between the visual scene and attention, grounding the semantic scene representation with word embeddings and evidencing the strong relationship between the semantic similarity of a scene region and the gaze-fixation pattern of the viewers.

Other researches on psycholinguistics have gauged the semantic distance of words via KB-based methods, as the latter provide a means to quantify the relationships between words and concepts within the semantic structure of a language. \cite{kenett2017semantic} created a Hebrew KB, quantifying semantic distance as the path length between word pairs by counting the number of steps for traversing from one word to another in the KB. They stated that a distance of 4 was the turning point for the performance in the semantic relatedness judgement task and subsequent recall from memory. Similarly, \cite{benedek2017semantic} proved that a KB based on semantic relatedness judgements was feasible and valuable to test the associative and executive accounts of creativity. Likewise, \cite{jelodar2010wordnet} proposed a model for concrete nouns' feature vectors based on Wordnet \citep{miller1995wordnet} similarity measurements for predicting brain activation patterns as measured with fMRI. It outperformed previous models.  

Psycholinguistic research on ageing is emerging, and in this context, embeddings and KBs are also gaining momentum in the study of ageing lexico-semantic processing \citep{wulff2019new}. For example, \cite{broderick2021dissociable} analysed the differences between the correspondence of two computational measurements (semantic dissimilarity and lexical surprisal) and event-related potentials in young and elderly subjects while performing a sentence comprehension task. They revealed dissociable neural correlates between the two populations, and showed that the two groups processed context-based predictions differently. \cite{cosgrove2021quantifying} also studied the impact of ageing on the performance of a verbal fluency task based on network sciences, and provided quantitative evidence of the decrease of flexibility in adults' semantic memory networks. 


\section{Extra linguistic features included in the dataset}
\label{sec:features}

Although the core of this dataset is word similarity calculation, the dataset has been controlled by four additional features: concreteness, frequency, semantic neighbourhood density and phonological neighbourhood density. The following sections summarize the scientific evidence highlighting these variables' significant role in psycholinguistics and NLP. A more profound comprehension of the connections between various linguistic features and word similarity from a cognitive viewpoint is needed. This dataset may lead to gaining insight into ongoing studies.

\subsection{Concreteness}
\label{ssec:cnc}
Concreteness is a term to refer to the degree to which a word denotes a tangible thing. This measurement was introduced by \cite{paivio1971imagery}, and it has been proven to play a role in several aspects, such as working memory \citep{mate2012you}, embodied cognition \citep{barsalou1999perceptual, fischer2008embodied, hauk2004somatotopic}, and neural representations on word processing \citep{wang2010neural}. The literature suggests that concrete words show thicker links to associated semantic information and involve visual imagery processes. Accordingly, they elicit faster response times and larger N400 and N700 electrophysiological signals \citep{ Schwanenflugel1991why,wang2010neural}. Remarkably, when contextual information and mental imageability are controlled, response times become faster for abstract words. However, the neural correlates do not change, suggesting that concrete words involve more significant semantic processing during meaning activation \citep{barber2013concreteness}.

Due to the relevance of concreteness in psycholinguistic research, several works on NLP have computationally predicted concreteness values using word embeddings \citep{ljubevsic2018predicting,charbonnier2019predicting,incitti2021fusing}, which have been proven to be useful in tasks like metaphor detection \citep{tsvetkov2014metaphor,alnafesah2020augmenting} and sentiment analysis \citep{rothe2016ultradense,long2019improving}. In \cite{long2019improving}, for example, the authors propose a cognition-grounded attention model in sentiment analysis, considering concreteness for leveraging the model’s attention mechanism. Further, concreteness seems to be gaining more attention in the NLP field, as some concreteness norms rated by humans have been published in various languages lately, being the English \citep{brysbaert2014concreteness} and Dutch \citep{brysbaert2014norms} ones quite extensive, and the Croatian \citep{coso2019affective}, Russian \citep{solovyev2022russian}, French \citep{bonin2018concreteness} and Spanish \citep{guasch2016spanish} ones more reduced.

\subsection{Word frequency}
\label{ssec:freq}

Word frequency is also key in both psycholinguistics and NLP. In the former, it is accepted that the rate at which something occurs in language affects how individuals process and recall information. This effect has been widely studied in lexical access \citep{balota1984lexical,balota2004visual,brysbaert2009moving} and recall tasks \citep{macleod1996word,gregg1976word,kinsbourne1974mechanism}, which involve memory usage. The effect of lexical frequency on human cognition has also been supported based on electrophysiological correlates \citep{strijkers2010tracking} and gaze-fixation patterns in spoken-word recognition \citep{dahan2001time} (for a review, see \cite{brysbaert2018word}.)

From a computational perspective, word frequency is also a fundamental feature in Text Meaning, and Information Retrieval tasks, among others. Language structure provides information about how important a word is in a text or corpus by measuring its occurrences. In NLP, it can be used in a wide variety of tasks, such as to determine the most frequent words in a language \citep{spink2001searching}, to identify rare words \citep{dave2003mining}, to synthesise information \citep{haghighi2009exploring} or to answer questions automatically \citep{koehn2017six}.

\subsection{Semantic Neighborhood Density}
\label{ssec:snd}

The semantic structure of the lexicon has always played an essential role in psycholinguistics and NLP, as the organization of knowledge is a pivotal aspect of studying meaning. Semantic neighbourhood density\footnote{Following the terminology in this field, we will refer to the size of the neighbourhood when using the term 'density', both for semantic and phonologic neighbourhoods.} (SND) of a word correlates with several cognitive processes and their respective brain activations but remains a poorly explored field. 

Different semantic neighbourhood size measures have been used in lexico-semantic research (e.g., metrics based on feature semantics, co-occurrence, and categorical relations), showing diverging effects of SND in lexical processing. For instance, when SND is measured considering semantic associations between words (or word co-occurrence is taken as a proxy of it), words with large semantic neighbourhoods generate faster responses than words with sparse semantic neighbourhoods in lexical decision tasks \citep{yates2003semantic, buchanan2001characterizing, locker2003semantic}. Opposite, when featural semantic information is taken as a ground to calculate the density, words with sparse neighbourhoods are related with faster recognition and naming \citep{rabovsky2016language, reilly2017effects}. In line with these findings, Abdel Rahman (2007) shows that associative relations facilitate word processing, whereas categorial connections between words create interference. Hence, different metrics of SND tap different constructs that need to be understood as complementary.

From an NLP perspective, KBs such as Wordnet \citep{miller1995wordnet}, FrameNet \citep{baker1998berkeley}, BabelNet \citep{navigli2010babelnet} or even Wikipedia provide a structured and quantifiable framework of words that show a strong correlation with our mental lexicon. Thus, they represent the primary computationally-grounded source for the semantic analysis of the neighbourhood of words via semantic relations.

\subsection{Phonological Neighborhood Density}
\label{ssec:phonn}

As with the SND, Phonological Neighborhood Density (PND) of words affects lexical processing, which indicates a need to control for such variables when designing a study. A word´s PND refers to the number of words in the lexicon that can be formed by substituting a single phoneme of the target word. Likewise, orthographic neighbourhood density (OND) is defined by the number of words that can be formed by replacing a single letter of the target word (Colheart et al.´s N metric) \citep{coltheart1977access}. 
These two measures imply considerable differences in opaque languages, such as English, Arabic or French, because phonemes do not present a one-to-one mapping into graphemes. Some studies on opaque languages have argued that neighbourhood effects reflect phonological rather than orthographic similarity \citep{Mulatti2006Neighborhood, yates2004theinfluence}; when orthographic similarity is controlled, phonological similarity still affects lexical decision times. However, PND and OND measures may be used indistinctly in shallow languages such as Basque and Spanish because there is a direct correspondence between phonemes and graphemes, with a few exceptions. Thus, in the present work, orthographic neighbours have been considered when computing the phonological neighbourhood density of noun pairs.

The influence of phonologically related words has been explored in different studies of phonological neighbourhood density, and it has shown relevant effects in task-dependent language processing. In particular, PND seems to exert competitive effects on word recognition tasks, but facilitatory effects in production tasks \citep{dell2011neighbors, gahl2012why}. In spoken word recognition, the acoustic stimulus activates potential candidates that are phonologically similar. Thus, the larger the phonological neighbourhood of a word, the harder it is to recognise the target word \citep{luce1998recognizing}. Paradoxically, in word production, dense phonological neighbours seem to ease production. For example, \citep{vitevitch2022theinfluence} found facilitatory effects of neighbourhood density in picture naming in a study that controlled other key factors such as frequency, neighbourhood frequency, familiarity and phonotactic probability. Although most studies have attested facilitatory effects in production, such effects are less consistent than in recognition tasks.

\section{Resources for the automatically created dataset}
\label{sec:dataset}

Two primary NLP sources of semantic information have been employed to create the Basque and Spanish semantic similarity datasets based on embeddings; KBs and text corpora. Regarding the former, the multilingual version of Wordnet \citep{miller1995wordnet} has been used due to its reliability as a KB in NLP and the abundance of associated libraries and tools. Table \ref{tab:reso} summarises the corpora and their derived embeddings which constitute the core of the resources in this work:

\begin{table}[h]
\centering
\begin{adjustbox}{width=1\textwidth}
\begin{tabular}{ll|l|l|}
\cline{3-4}
                                &            & Acronym                  & Description                               \\ \hline
\multicolumn{1}{|l|}{}          & Corpora    & txt                      & Text corpora                              \\
\multicolumn{1}{|l|}{}          &            & kb                       & Wordnet-based pseudo-corpora              \\ \cline{2-4} 
\multicolumn{1}{|l|}{}          & embeddings & FT\textsubscript{txt}    & Fastext over text corpora                 \\
\multicolumn{1}{|l|}{}          &            & FT\textsubscript{kb}     & Fastext over wordnet-based pseudo-corpora \\
\multicolumn{1}{|l|}{}          &            & FT\textsubscript{hyb}   & Combination of previous two embeddings    \\ \hline
\end{tabular}
\end{adjustbox}
\caption{Acronyms along with the description of the main resources used in this work. The first group includes two types of corpora, KB and text corpora (\emph{kb} and \emph{txt}, respectively). The second group describes the types of embeddings, namely FasText over text corpora, over wordnet pseudo-corpora and the meta-embeddings, which combine the first two types 
(\emph{TF\textsubscript{txt}}, \emph{TF\textsubscript{kb}} and \emph{FT\textsubscript{hyb}}, respectively).}
\label{tab:reso}
\end{table}

\subsection{Wordnets and NLTK toolkit}
\label{sec:wn}

Wordnet \citep{miller1995wordnet} is an English lexical database organized by concept and meaning. Specifically, lexical forms of nouns, verbs, adjectives and adverbs are grouped into sets of cognitive synonyms called synsets, each expressing a language-independent concept. Further, each synset is linked to several synsets via semantic relations such as hypernymy, hyponymy, meronymy and antonymy, thus, creating a semantic network.


The semantic structure of Wordnet gives us a robust framework for the present work. First, the nature of the semantic information it conveys around words and concepts allows assigning two of the previously mentioned features to each word of the dataset; SND and concreteness. Second, the semantic structure of the WordNet grants coding in two of the three types of embeddings employed in this work, namely wordnet and hybrid embeddings (see \ref{sec:emb}).

We have used the Open Multilingual Wordnet (OMW)\footnote{\url{http://compling.hss.ntu.edu.sg/omw/}} which is linked to the Princeton WordNet 3.0 (PWN)\footnote{\url{https://wordnet.princeton.edu/}}. OMW extends the former English Wordnet to 34 languages, including Basque and Spanish, and uses the PWN synset structure. Thus, it maps the lexicalizations of all languages into the same semantic network. This extension method based on English Wordnet's structure is called \emph{expand-approach}\footnote{In this paper, we will refer to the original PWN and the one used in this work (Wordnet 3.0) as 'Wordnet'. In contrast, we will use 'wordnet' for the rest which derive from PWM such as the Spanish and Basque ones.}. We have used Python NLTK toolkit\footnote{\url{https://www.nltk.org/howto/wordnet.html}} to extract concreteness and SND features from Wordnet synsets (see section \ref{sec:feat}), for both Basque and Spanish.

In parallel to the KB Wordnet, we have used language corpora as a complementary semantic information resource, which is introduced in the next section.

\subsection{Corpora}
\label{sec:cor}

In the following section, we describe the two types of corpora used during the construction of the dataset; the text corpora and the wordnet-based KB pseudo-corpora.

\subsubsection{Text corpora}
\label{sec:txtcor}

Text corpora are collections of vast amounts of texts, being usually annotated (e.g., British National Corpus) \footnote{\url{https://www.english-corpora.org/bnc/}} or even syntactically parsed (e.g., Penn TreeBank \citep{marcus1993building}). Plain text corpora with no structure have been used to create this dataset.  The use of plain text has allowed using greater text corpora in the computation of the relationships of words and their contexts via word occurrence. Hence, we have calculated directly from the corpora the similarity between words by using word embeddings. In the case of Basque, it has also been used for extracting word frequencies (see section \ref{sec:featfreq}).

In Spanish, we have employed the readily available 2018 Wikipedia dump text corpora\footnote{https://linguatools.org/tools/corpora/wikipedia-monolingual-corpora/}, extracting the text from the dump using a script\footnote{\texttt{xml2textx} available in the same site. The content of tables and maths have been deleted.}. Since the size of Wikipedia is insufficient for building a large corpus in low-resourced languages such as Basque, web crawling has been used to complement the corpora in this language, since it is an effective method for collecting texts to compensate for this deficiency \citep{leturia2012evaluating}. Thus for the case of Basque, we have employed the publicly available Euscrawl corpus\footnote{http://ixa.ehu.eus/euscrawl/} \citep{artetxe2022does}. 

Both Basque and Spanish corpora were pre-processed with the standard procedure, that is, lowercase and tokenization. Token sparsity in Basque, for being an agglutinative language, was avoided with an in-house stemmer. The final Spanish and Basque corpus comprises 608 and 288 million tokens, respectively.  


\subsubsection{Knowledge-based corpora}
\label{sec:kbcor}

This kind of corpora is not as known as the text corpora, but it has recently gained some attention in the NLP field. The knowledge-based corpora latently contain the semantic structure of a KB (in our case Basque or Spanish wordnets), and we have dubbed it as pseudo-corpora in the present work because it is not human-understandable. This pseudo-corpora represents concepts and lexicalizations of knowledge bases in a much more compact format than traditional methods. As explained in the following sections (see \ref{sec:emb}), the pseudo-corpora have been processed by a neural network model to encode the semantic structure of Basque or Spanish wordnets in a continuous vector space.

To do this, we applied the monolingual method for English introduced by \cite{goikoetxea2015random}, but in the Basque and Spanish settings. This technique uses a Monte Carlo method to compute the PageRank algorithm \cite{avrachenkov2007monte}. The algorithm considers the KB as an undirected graph comprised of concepts and links among concepts. It needs a dictionary which associates words with concepts, as well as a damping factor $\alpha$ that determines the continuity of the random walk and the maximum number of walks.

For creating every context \footnote{Context refers to each of the lines in the pseudo-corpora created by the algorithm.} the algorithm starts in a random concept and launches a random lexical form of the concept via the dictionary. Afterwards, it decides whether to jump to another concept\footnote{This will depend on the parameter $\alpha$} and to launch a random lexical form of the latter, or stops the walk and starts over a new walk. Finally, if the number of walks reaches the maximum number of walks, it ends the process. Note that the word is fed to a text file whenever the method launches a lexical form in the walk. 

Each line of the following example shows a different walk of the monolingual algorithm in Wordnet 3.0, which is used by \cite{goikoetxea2015random}. In this case, every walk has a different length, and each jump from concept to concept has launched a random lexicalization. It is worth mentioning that this pseudo-corpus is not human-readable, but every walk gathers semantically related words, so that implicitly it contains Wordnet's structure. The following example shows five random walks from an English Wordnet pseudo-corpus:

\begin{center}
\begin{displayquote}
    \begin{verbatim}
        storyteller liar beguiler grifter dissimulation
        revitalize strength delicate ethereal
        paved patio terrace house living_room home
        swimming dive
        backlog fire re-afforest forest woods rainforest
     \end{verbatim}
\end{displayquote}
\end{center}


As mentioned before, the monolingual version of this method is adapted to the Basque and Spanish setup, using the dictionaries of both languages from OMW; hence, aligned with Wordnet 3.0 semantic structure. This means that the former English lexical forms have been translated to the target language, but the semantic structure remains intact. The resultant wordnets' size is not the same for every language, since \textit{expand-approach} wordnets do not have the same number of lexicalizations as the former English Wordnet 3.0. As shown in Table \ref{tab_wn}, Basque and Spanish wordnets' sizes are much smaller than the original one.

    \begin{table}[h]
        \centering
        \begin{tabular}{l|cc|}
        \cline{2-3}
                                  & Lexicalizations & Synsets \\ \hline
        \multicolumn{1}{|l|}{EN}  & 147306          & 136334  \\
        \multicolumn{1}{|l|}{ES}  & 53039           & 55814   \\
        \multicolumn{1}{|l|}{EU}  & 26701           & 30464   \\ \hline
        \end{tabular}
        \centering
        \caption{Wordnet sizes for English (EN), Spanish (ES)
        and Basque (EU). Number of lexicalizations and synsets
        in the middle and rightmost columns, respectively.}
        \label{tab_wn}
    \end{table}

The reduction of the size of the semantic structure and the number of lexical forms in Basque and Spanish directly impacts the dimension of the wordnet-based corpora in both languages. To prevent saturation and redundant information as described by \citep{goikoetxea2015random,goikoetxea2018bilingual}, the sizes of the pseudo-corpora in both languages (see section \ref{sec:cor}) have been limited. Even though the wordnet-based corpus is smaller in Basque, they are big enough to encode their respective wordnets' semantic structure (see section \ref{sec:emb}). 

English Wordnet 3.0 with glosses consists of 147306 lexical forms, and \cite{goikoetxea2016single} reached the peak performance in word similarity task with 200 million random walks, which created an 1100 million token corpus. In this work, we kept the same proportion between the number of lexical forms and random walks as \cite{goikoetxea2016single} in order to ensure good performance. Thus 72 million and 36.3 million random walks were performed for Spanish and Basque, respectively, resulting in a corpus of 406 million tokens for the former and 166 million for the latter.

\subsection{Embeddings}
\label{sec:emb}

Three types of static embeddings have been computed: text embeddings ($FT_{txt}$), wordnet-based embeddings ($FT_{kb}$), and hybrid embedding ($FT_{hyb}$). These word representations encompass the two sources of semantic information (text and KB embeddings) related to word similarity (see section \ref{sec:relwork}).

The text embeddings are low-dimensional word representations computed following the Distributional Hypothesis \cite{harris1954distributional}. More precisely, a neural network processes the whole text corpus by traversing the corpus word by word, computing words' representations based on the Distributional Hypothesis. It calculates the representation of a given word based on the representations of all the neighbours found within a predefined window while traversing the corpus. Eventually, every token in the corpus ends up with a vector representation (embedding) that latently encodes its semantic (and syntactic) features.

In this work, the neural-based model \texttt{fastText} \citep{bojanowski2017enriching} has been used because of its more robust performance. It implements a variant of the Distributional Hypothesis which, instead of exploiting the neighbouring words' information to compute a representation of a given word, exploits subword information. Both text-based and wordnet-based embeddings have been calculated using \texttt{fastText}.

In the case of Basque and Spanish text-based embeddings, we fed \texttt{fastText} with their respective text corpora (see \ref{sec:cor}) separately. Regarding wordnet-based ones, we encoded the semantic structure of Basque and Spanish wordnets in a vector space following the method proposed by \citep{goikoetxea2015random}, which comprises two steps. First, the creation of Basque and Spanish Wordnet pseudo-corpus, as explained in the previous section. Second, processing of Basque and Spanish pseudo-corpora with a neural-based model to obtain their respective wordnet embeddings. In the original proposal, \citep{goikoetxea2015random} used \texttt{word2vec} \citep{mikolov2013distributed}, but as mentioned before, in the present work \texttt{fastText} has been used with the same parameters as in \cite{mikolov2018advances}.

Recent works show that embeddings which combine semantic information from both text and KB (i.e., hybrid embeddings) have an overall higher performance in word similarity tasks. \cite{lastra2019reproducible} proved that hybrid embeddings outperformed most ontology-based measures and the rest of the word embedding models. Likewise, \citep{goikoetxea2016single, garcia2020common} proved that hybrid embeddings are more explicative of human perception of semantic distance than text or KB embeddings separately. In the present work, the method proposed originally by \cite{goikoetxea2016single} and improved by \cite{goikoetxea2016single} and \cite{garcia2020common} will be implemented in the computation of hybrid embeddings. In short, \citep{garcia2020common} proposal consists in four steps\footnote{For further details, see \cite{garcia2020common}.}:

\begin{enumerate}
    \item To compute separate text and wordnet embeddings.
    \item To map text embeddings space onto the wordnet one.
    \item To estimate word embeddings for both spaces.
    \item To combine equivalent text and wordnet embeddings.
\end{enumerate}

In the creation of $FT_{hyb}$ embeddings, \cite{garcia2020common} have been strictly followed, employing \texttt{vecmap} \citep{artetxe2018acl} for the mapping of text and wordnet embeddings.

\subsubsection{Evaluation of the quality of the embeddings in similarity task}
\label{sec:evalemb}

To verify the quality of the three types of embeddings, we have tested them in a word similarity task. In NLP, word similarity is commonly calculated by the cosine similarity of the angle between two-word embeddings, measuring the similarity of the words they represent. The cosine similarity is determined by computing the dot product of the two vectors and dividing it by the product of the Euclidean norms of the vectors. The cosine similarity is a measure ranging from 0 to 1, where 0 means the complete absence of similarity and 1 means complete similarity (i.e., synonyms). The similarity measurement is independent of the origins of the embeddings, in the way that that text, wordnet and hybrid representations are processed in the same way.

In this paper, the term similarity has been used indistinctly, but the difference between pure similarity and relatedness has been widely recognised in cognitive sciences for a long time \citep{tversky1977features}. Pure similarity measures the degree to which two concepts share semantic features, while relatedness is the degree of association between two words. Regarding the semantic relations involved, pure similarity includes synonymy and hyponymy/hyperonymy. In contrast, relatedness encompasses the previous ones and a wider variety of relations, such as meronymy, functional associations and other unusual relations. For example, \textit{wolf} and \textit{dog} are taxonomically linked by hypernymy relations in the same semantic structure and share many features; thus, they have high similarity. In contrast, \textit{wolf} and \textit{moon} do not share any semantic feature (low similarity) but are related by association; hence, they have a high relatedness.

Cosine similarity measurement does not distinguish between these two aspects but can be applied to different types of embeddings to measure distinctive semantic relations. In this work, we have chosen three types of embeddings, which perform differently in pure semantic similarity \citep{hill2015simlex,agirre2009study,rubenstein1965contextual} and relatedness \citep{finkelstein2001placing,bruni2014multimodal} measures. Broadly, wordnet embeddings are more precise for measuring pure similarity relations, text embeddings are more sensitive to compute relatedness, and hybrid embeddings have been proven to be more robust for capturing semantic relations in general \citep{goikoetxea2016single,goikoetxea2018bilingual, garcia2020common}. 

Hence in this work, we chose the available similarity and relatedness datasets for evaluating all embeddings in both languages. In Spanish, we operated with the pure similarity datasets RG65 (RG) \citep{camacho2015framework} and SimLex999 (SL) \citep{etcheverry2016spanish}, and the relatedness dataset Wordsim353 (WS) \citep{hassan2009cross}. In the case of Basque, we used the pure similarity RG dataset and the relatedness dataset WS created by \cite{goikoetxea2018bilingual}. 

In both Spanish and Basque, $FT_{txt}$, $FT_{kb}$ and $FT_{hyb}$ representations have been compared with the baseline, using publicly available text-based representations \texttt{fastText} \footnote{https://fasttext.cc/docs/en/crawl-vectors.html}. All of our embeddings have been computed with the set of parameters as in the baseline embeddings. The Spearman correlation results between the embeddings and the gold standards are shown in table \ref{tab:emb} and discussed below:



\begin{table}[h]
\centering
\begin{tabular}{ll|lll|}
\cline{3-5}
                                         &      & \multicolumn{3}{c|}{Dataset}                                                                       \\
\multicolumn{1}{c}{\textbf{}}            &      & \multicolumn{1}{c}{RG}             & \multicolumn{1}{c}{WS}             & \multicolumn{1}{c|}{SL}  \\ \hline
\multicolumn{1}{|l}{\multirow{4}{*}{EU}} & Baseline  & \multicolumn{1}{c}{77.05}          & \multicolumn{1}{c}{65.7}           & \multicolumn{1}{c|}{---} \\
\multicolumn{1}{|l}{}                    & $FT_{txt}$  & 77.86                              & 73.31                              & \multicolumn{1}{c|}{---} \\
\multicolumn{1}{|l}{}                    & $FT_{kb}$   & 85.67                              & 65.88                              & \multicolumn{1}{c|}{---} \\
\multicolumn{1}{|l}{}                    & $FT_{hyb}$  & \multicolumn{1}{c}{\textbf{86.55}} & \multicolumn{1}{c}{\textbf{74.57}} & \multicolumn{1}{c|}{---} \\ \hline
\multicolumn{1}{|l}{\multirow{4}{*}{ES}} & Baseline & \textbf{87.9}                 & 57.8                &  36.58                   \\
\multicolumn{1}{|l}{}                    & $FT_{txt}$  & 86.57                              & 57.28                              & 28.7                    \\
\multicolumn{1}{|l}{}                    & $FT_{kb}$   & 72.84                              & 57.32                              & 39.93                    \\
\multicolumn{1}{|l}{}                    & $FT_{hyb}$  & 87.25                             & \textbf{63.45}                      & \textbf{40.57}                         \\ \hline
\end{tabular}
 \caption{Word similarity task results in RG, Wordsim353 (WS) and SimLex999 (SL) datasets in Basque (EU) and Spanish (ES). Text ($FT_{txt}$), wordnet-based ($FT_{kb}$) and hybrid embedding ($FT_{hyb}$) representations are compared to their baselines. The best results for each dataset and language are expressed in bold.}
    \label{tab:emb}
\end{table}

\begin{itemize}
    \item $FT_{txt}$ embeddings: $FT_{txt}$ results perform similarly to the baseline ones in both languages, with two exceptions. One, the Basque $FT_{txt}$ result in the WS dataset is higher than the baseline, likely due to the greater corpus size Euscrawl. Second, the Spanish $FT_{txt}$ in the SL dataset is lower than the baseline. The only plausible explanation for this result in the Spanish SL may lie in the differences in the pre-processing of the corpus; the baseline text corpus was tokenized with Europarl pre-processing tools\footnote{\url{https://www.statmt.org/europarl/}}, whereas the NLTK tokenizer has been used in the present work.
    
    \item $FT_{kb}$ embeddings: being this type of embeddings more suited for pure similarity datasets (RG and SL) rather than for relatedness ones (WS), $FT_{kb}$ results are higher than the baseline in the Basque RG and the Spanish SL, but not in the Spanish RG. This underperformance in Spanish must be interpreted with caution, as the RG dataset is small (64 pairs) and it has low statistical power. 
    
    The most noticeable result, though, is in the relatedness WS dataset, in which $FT_{kb}$ performs at baseline. Note that the baseline txt-based embeddings are supposed to have a better performance in WS due to their capacity of better capturing relatedness relations, so we expect them to obtain better results than the $FT_{kb}$ measurement. However, the incorporation of gloss relations (which are relatedness relations) when creating the wordnet-based corpora may has enhanced the capability of $FT_{kb}$ embeddings to measure relatedness.
      
    \item $FT_{hyb}$ embeddings: comparing the results of $FT_{txt}$, $FT_{kb}$ and $FT_{hyb}$ embeddings, we find that the combination of the first two into the latter enhances their performance in similarity tasks for both languages across all datasets. $FT_{hyb}$ also outperforms the baseline in all datasets, with the only exception of RG in Spanish, likely due to its small size, as mentioned above. 
    
\end{itemize}

In sum, the three types of embeddings in this work have performed as expected, and furthermore, $FT_{hyb}$ embeddings have shown overall best results.  Moreover, the quality of the embeddings in the Basque language has been proven to be the best known to date. Note that the excellence of the embeddings is critical in this work, as they are used to create the three semantic similarity measurements (see section \ref{sec:sim}) that constitute the core features of the dataset. Altogether, all embeddings created in this work have been considered suitable to be included in the final dataset.


\section{Building the dataset}
\label{sec:createdata}



This work has aimed to create a dataset of noun pairsthat compiles information about the semantic similarity between them. As already introduced, the semantic similarity has been calculated using three types of embeddings; text, wordnet and hybrid embeddings. The linguistic features controlled in pairing the nouns have been concreteness, word frequency, and semantic (SND) and phonological (PND) neighbour density. 

Each feature has been clustered via KNN classification \citep{cover1967nearest} to match pairs of words. That is, each pair of nouns has been set every time there has been a coincidence in the four feature clusters. Despite maximizing the number of potential noun pairs, a two-group clustering (low and high values) has resulted balanced in concreteness, word frequency, SND and PND (see Table \ref{tab:dictclus}).

Each line of the final dataset comprises the normalised values of features of the noun pairs, their corresponding clusters and the three similarity measurements. Each language has a separate dataset built by following the next steps: 

\begin{itemize}
    \item First, compute the four features for every noun in the corresponding Basque or Spanish wordnet.
    \item Second, find every possible noun pairs thatmatch the four linguistic features via clustering.
    \item Third, compute similarities for every noun pair.
    \item Last, write every noun pair's three similarity measurements along with their features' values (L2-normalised) and cluster number.
\end{itemize}

The following sections are dedicated to describing the details of the measurements of every feature mentioned in the above paragraphs and the creation of the final dataset.

\subsection{Similarity measurements}
\label{sec:sim}

Table \ref{tab:simvals} shows the percentages of semantic similarity values for all possible noun pairs classified along five different ranks. Only nouns with the three types of embeddings that match all four linguistic features are considered. The percentages presented in table \ref{tab:simvals} show a bias towards lower rank similarity values for all types of embeddings in both Basque and Spanish. This is even more evident in the wordnet embeddings.

This phenomenon was already pointed out in a cross-lingual setting by \cite{lample2019cross}. The author observed that \url{fasText} based embeddings' cosine similarity mean value was considerably lower than that achieved by XLM \citep{lample2019cross}, a large cross-lingual language model. Authors suggested that the nature of these two types of vector spaces may result in higher word similarity values, because language models' vector spaces are more compact than the \url{fasText} due to the formers training in a sentence encoder. This phenomenon only shows that the non-contextual embeddings like \url{fasText} or \url{word2vec} are sparser than the language model ones. As seen in the results of table \ref{tab:emb}, the different organization of the lexicon in the vector spaces does not affect the performance of the embeddings in the word similarity task.

\begin{table}[]
\centering
\begin{tabular}{l|ccc|ccl|}
\cline{2-7}
                              & \multicolumn{3}{c|}{\cellcolor[HTML]{C0C0C0}EUS}                   & \multicolumn{3}{c|}{\cellcolor[HTML]{C0C0C0}ESP}                       \\
                              & \cellcolor[HTML]{FFFFFF}$FT_{hyb}$ & \cellcolor[HTML]{FFFFFF}$FT_{kb}$ & $FT_{txt}$     & $FT_{hyb}$                  & $FT_{kb}$                   & \multicolumn{1}{c|}{$FT_{txt}$} \\ \hline
\multicolumn{1}{|c|}{0.0-0.2} & 89.1        & 98.44   & 84.9    &  93.52       &   98.68     &    89.13      \\
\multicolumn{1}{|c|}{0.2-0.4} & 10,3        & 1.48    & 13.97   &  6.09       &   1.24      &    10.12      \\
\multicolumn{1}{|l|}{0.4-0.6} & 0.54        & 0.066   & 1.045   &  0.36       &   0.06     &    0.71       \\
\multicolumn{1}{|l|}{0.6-0.8} & 0.025       & 0.009  & 0,048  &  0.015       &   0.007   &    0.03       \\
\multicolumn{1}{|l|}{0.8-1.0} & 0.00135     & 0.0015  & 0.0018 &  0.001       &   0.0013    &    0.0014    \\ \hline
\end{tabular}
    \caption{ Percentages of word similarity values in the Basque and Spanish dataset across five ranges and three types of embeddings; text ($FT_{txt}$ ), wordnet ($FT_{kb}$ ), and hybrid ($FT_{hyb}$ ).}
    \label{tab:simvals}
\end{table}

\subsection{Features}
\label{sec:feat}    

This section describes the details of the calculation of each linguistic feature before being L2-normalised and clustered in the final dataset.


\subsubsection{Concreteness}
\label{sec:featcon}


Concreteness measurements were calculated automatically by exploiting Wordnet's taxonomy in Basque and Spanish. For doing so, \cite{feng2011simulating} was followed, who proposed an algorithm for predicting word concreteness via various lexical resources and features, namely, human ratings, lexical types, latent semantic analysis dimensions, word frequency and length, and hypernymy level. Among these several aspects, in this work, only the \emph{Hypernymy Level} was considered for computing concreteness. This is because even though human ratings and lexical type are not included in this research, the rest are going to be analysed and treated as separate features. \footnote{Although latent semantic analysis is not used for word representations in the present word, more advanced techniques of static word embeddings are included (see section \ref{sec:emb})}.
The \emph{Hypernymy Level} aspect scores the concreteness of a word following its hypernymy relations in Wordnet's hierarchical structure. For every noun in Basque and Spanish, the mentioned method does the following in the corresponding wordnet:\vspace{-0.3cm}
\begin{itemize}
    \item Check all of its synsets within a given noun.
    \vspace{-0.3cm}
    \item For each synset, the method counts every hypernym from the source synset until the topmost synset (i.e., Wordnet's root node \emph{entity}), thus scoring the depth of the source synset in Wordnet's tree structure.
    \vspace{-0.3cm}
    \item Compute the concreteness of a given noun by averaging all the depths of its synsets.
\end{itemize}

Note that the higher the depth, the more concrete the word is, and the lower the depth more abstract it will be. For example, the word \emph{car} has five synsets in Wordnet 3.0. If we choose the most common synset for that noun\footnote{The sense with the following definition in Wordnet: \emph{a motor vehicle with four wheels; usually propelled by an internal combustion engine.}}, its hypernymy path is the following: 

$$\textbf{car}  \rightarrow \emph{MotorVehicle}  \rightarrow \emph{SelfPropelledVehicle}  \rightarrow \emph{WheeledVehicle}  \rightarrow \emph{Container} $$
$$\rightarrow \emph{Instrumentality}  \rightarrow \emph{Artifact} \rightarrow \emph{Whole} \rightarrow \emph{Object} \rightarrow \emph{PhysicalEntity}  \rightarrow \textbf{Entity}$$. 

Thus, the depth of the path of the mentioned synset of \emph{car} is 10. The word \emph{Wheeled Vehicle} has a depth of 7, meaning that it is less concrete than \emph{car} and \emph{ambulance} a depth of 11, therefore being more concrete than \emph{car}. For the final score of the noun, the method computes all the depths of the rest of the synsets and averages all of them.

    
\subsubsection{Semantic neighbourhood density}
\label{sec:featsemn}

This work's framework for computing SND is also based on Wordnet's semantic structure. So, the semantic neighbours of every noun are computed by counting all semantic relations of its synsets. A procedure similar to the one described in section \ref{sec:featcon} has been used for every noun in Basque and Spanish:

\vspace{-0.3cm}
\begin{itemize}
    \item Check all of its synsets within a given noun.
    \vspace{-0.3cm}
    \item For each synset, count all its surrounding synsets linked by a Wordnet semantic relation.
    This approach departs from the source synset and checks all its semantic relations to find its neighbouring synsets.
    \vspace{-0.3cm}
    \item Finally, compute the semantic neighbours of a noun averaging all the counts in its synsets.
\end{itemize}

Only first-degree semantic neighbours have been considered. Regarding the Wordnet semantic relations, we have taken every available synset-based semantic relation in the NLTK toolkit and discarded the more surface-level lexical relations like derivation and pertainymy. 

Following the same example as in the previous section, the most common synset for the noun \emph{car} has 31 hyponyms, 29 meronym parts and one hypernym. That is, 61 first-degree semantic neighbours.


\subsubsection{Phonological neighbourhood density}
\label{sec:featphon}

Phonological neighbours are pairs of words that differ in only one phonological segment, such as \textit{cat} and \textit{bat}.  For this work, phonological neighbourhood size has been calculated with the Levenstein distance method \cite{levenshtein1965binary}. It has several advantages compared to other alternatives, such as the Hamming Distance or the Jaro-Winkler distance; it accounts for both substitution and insertion/deletion operations, so the distance is more accurately computed, independent of the length of the strings. 

This work assumes that two words are phonological neighbours if their Levenstein distance is smaller or equal to one. For example, in the case of the word \emph{car}, the number of phonological neighbours that meet the conditions mentioned above is 31, with first-degree neighbouring words such as \emph{cat}, \emph{card}, \emph{scar}, \emph{ear}, \emph{jar} and \emph{tar}. Multiword expressions and nouns with less than three characters in length have been excluded from the dataset.





\subsubsection{Word frequency}
\label{sec:featfreq}

Word frequency measures have been calculated using Zipf frequencies; a base-10 logarithm of its occurrences per billion words. Zipf frequency allows for a more comprehensive analysis than raw noun counts because it accounts for the relative importance of a word in the corpus. 

For Spanish, Python's \url{wordfreq} library has been used to obtain Zipf frequencies. Given the nature of the 10-base logarithm scale, Zipf values in this library are within the range of 0-8. In Basque, Zipf frequencies have been computed from the most extensive Basque public corpus, Euscrawl \citep{artetxe2022does}. Euscrawl corpus has been pre-processed as described in \ref{sec:cor}, and raw word counts have been calculated using \url{fastext} model from the \url{gemsim} library in Python. Afterwards, those counts were converted into Zipf frequencies, limiting their maximum and minimum values between 0 and 8.



\subsubsection{Feature dictionaries}
\label{sec:featdict}

As mentioned in previous sections, each noun is accompanied by four features in the dataset. An independent dictionary has been created for each feature (i.e., concreteness, frequency, SND and PND) in Basque and Spanish, so that it assigns the measurements of its feature along with its cluster number to each noun. Once the feature dictionaries have been computed, the next step has been building the final dataset (see section \ref{sec:constmatrix}).

The lists of single-word nouns from Spanish and Basque wordnets have been extracted via NLTK toolkit, constraining the candidates to 28647 and 22877 in each language, respectively. 

Although the Spanish wordnet nearly doubles the Basque one in size (see Table \ref{tab_wn}), the unbalance is not pronounced in the size of the feature dictionaries because the number of available nouns in the NLTK toolkit is comparable across these two languages. To improve the KNN classification of the nouns across each feature, outliers have been removed using the interquartile range, and L2-normalisation has been applied to every raw measurement. This adjustment has diminished the size of the datasets but has reduced noisy samples and balanced cluster sizes.

Table \ref{tab:dictsize} shows the resulting feature dictionaries' size and the mean value for each feature.

\begin{table}[h]
\centering
\begin{tabular}{ll|ll|}
\cline{3-4}
                                             &     & \multicolumn{1}{c}{\cellcolor[HTML]{C0C0C0}EU} & \multicolumn{1}{c}{\cellcolor[HTML]{C0C0C0}ES} \\ \hline
\multicolumn{1}{|c}{}                        & size & 19660                       & 14771                       \\
\multicolumn{1}{|c}{\multirow{-2}{*}{CNC}} & avg &  8.41                       & 8.38                            \\ \hline
\multicolumn{1}{|c}{}                        & size & 14380                       & 12146                       \\
\multicolumn{1}{|c}{\multirow{-2}{*}{FRQ}} & avg &  6.39                       & 2.95                             \\ \hline
\multicolumn{1}{|c}{}                        & size & 18044                       & 15534                       \\
\multicolumn{1}{|c}{\multirow{-2}{*}{SND}}  & acg &  2.47                       &  2.18                           \\ \hline
\multicolumn{1}{|c}{}                        & size & 14671                       & 14608                       \\
\multicolumn{1}{|c}{\multirow{-2}{*}{PND}} & avg &  1.75                       &  1.38                           \\ \hline
\end{tabular}
\caption{Concreteness (CNC), frequency (FRQ), phonological neighbourhood density (PND) and semantic neighbourhood density (SND) dictionaries for Basque (EU) and Spanish (ES). The upper line in each feature shows the size of a specific dictionary, and the lower line shows the mean value of each measurement.}
\label{tab:dictsize}
\end{table}

Table \ref{tab:dictsize} shows that the Basque feature dictionaries are slightly bigger than the ones in Spanish, likely because there are more outliers in the Spanish dataset. Regarding the average value of each feature, the only difference is in frequency. On average, Basque nouns are more frequent than the Spanish ones\footnote{Note that the mean Zipf frequency value of the Basque words is 6, meaning that a word appears once per a thousand words, whereas the mean Zipf value in Spanish is 2, indicating that a word appears once per million words.}. A reasonable explanation for this difference may be the size of the corpora used to calculate word frequencies in each language. The Scrawl corpus is smaller than the one used for computing Spanish Zipf frequencies in the NLTK toolkit, and it has much fewer infrequent words. This may bias the Zipf frequency towards a higher average value in Basque.

    


Table \ref{tab:dictclus} shows the token distribution of all previous dictionaries in KNN-based clusters. Features are clustered in two groups, differencing between high and low values. The size of the dictionary of phonological neighbours features is far below the rest. This is because nouns with 0 neighbours have been excluded from the KNN classification (14671 in Basque and 12085 in Spanish) because they tend to unbalance the cluster distribution, leaving the cluster with high-PND value with almost no content. Therefore, nouns with 0 neighbours have been extracted into a subgroup and treated as a separate cluster when creating the final dataset.

\begin{table}[]
\centering
\begin{tabular}{l|cc|cc|}
\cline{2-5}
                            & \multicolumn{2}{c|}{\cellcolor[HTML]{C0C0C0}EUS}      & \multicolumn{2}{c|}{\cellcolor[HTML]{C0C0C0}ESP} \\
                            & \cellcolor[HTML]{FFFFFF}L & \cellcolor[HTML]{FFFFFF}H & L                       & H                      \\ \hline
\multicolumn{1}{|l|}{CNC} & 11373                     & 8287                      & 9692                    & 6179                   \\
\multicolumn{1}{|l|}{FRQ} & 5646                      & 8734                      & 6103                    & 6043                   \\
\multicolumn{1}{|l|}{SND}  & 13915                     & 4129                      & 4772                    & 9836                   \\
\multicolumn{1}{|l|}{PND} & 3791                      & 974                       & 1017                    & 2432                   \\ \hline
\end{tabular}
\caption{Concreteness (CNC), frequency (FRQ), phonological neighbourhood density (PND) and semantic neighbourhood density (SND) dictionaries for Basque (EU) and Spanish (ES). The upper line in each feature shows the size of a specific dictionary, and the lower line shows the mean value of each measurement}
\label{tab:dictclus}
\end{table}



\subsection{Construction of word pair matrix}
\label{sec:constmatrix}

The final step in this work has consisted of creating the noun pair dataset out of the feature dictionaries described previously. There are two conditions to fulfil to set a pair of nouns: 

\begin{itemize}
    \item First, the two nouns composing the pair must be represented in the three types of embeddings.
    \vspace{-0.3cm}
    \item Second, the two nouns composing the pair mustshare cluster in all four linguistic features.
\end{itemize}

The number of nouns that have fulfilled the first condition was more considerable in Basque (n= 15110) than in Spanish (n= 8565). This phenomenon is due to the high number of multiword expressions in the Spanish wordnet, which are discharged from the dataset. The suppression of the multiword leads to a lower overlap of text and wordnet embeddings and, therefore, a lower amount of hybrid embeddings.

All the nouns that have fulfilled the first condition have been traversed to find all possible pairs of nouns that match the four linguistic features, as stated in the second condition. Every time a pair has been set, featural values in concreteness, frequency, SND, and PND, as well as their cluster identification\footnote{We used -1 number for 0 neighbour subgroup in phonological neighbour matching, as that group was not part of a KNN cluster. For the rest of the linguistic features, clustering was marked with 0 and 1 to indicate low or high-value clustering}, have been inserted in the dataset, along with the three types of similarity measurements. Altogether, each nour pair contains pieces of information spread over 15 columns.




\section{Conclusions}
\label{sec:conclusions}

Psycholinguistic evidence supports the idea that the overlap of semantic features across concepts is crucial in the computation of semantic relations, and by extension, in semantic processing. A comprehensive work on NLP has shown that embeddings or vectors are a sensitive method to artificially compute similarities across concepts by accounting for large number of semantic properties, which may be applied to different types of linguistic resources in order to uncover distinctive semantic relations.  This work has aimed to bridge computational linguistics and psycholinguistics to automatically build a dataset of vectorised word similarity measures in Basque and Spanish\footnote{Datasets will be available upon acceptance}.

We have presented a computationally grounded dataset that encompasses different aspects of the semantic information present in both text corpora and knowledge bases. On the one hand, each dataset includes three similarity measurements based on their corresponding embedding computations, namely, text-based, wordnet-based and hybrid embeddings. These measurements encode different subtleties of meaning. Text-based embeddings are computed out of word co-occurrence in large natural language corpora; being prone to better measure relatedness relations. Wordnet-based embeddings encode the semantic structure of Wordnet, so that they are considered a measurement of pure similarity relations. Finally, hybrid embeddings combine both text-based and wordnet-based embeddings, binding categorical and associative relations. In addition, all the materials have been controlled for several linguistic features (concreteness, frequency and semantic and phonemic neighbour size) to adjust the dataset to various research interests and requirements.

\clearpage

\section{Acknowledgements}
\label{sec:ack}
This work was supported by the Leonardo grant of the BBVA Foundation for Researchers and Cultural Creators 2021 (FP2157) (J.G., M.A., I.S.), Consolidated Group funding from the Basque Government (IT1439/22-GIC21/132) (M.A., I.S.), and the grant RYC2021-033222-I funded by the Ministry of Science and Innovation/State Research Agency/ 10.13039/501100011033 and by the European Union NextGenerationEU/Recovery, Transformation and Resilience Plan of Spain (M.A.).

\clearpage

\bibliographystyle{elsarticle-harv}
\bibliography{ms}

\clearpage

\end{document}